\documentclass{article}
\usepackage{iclr2026_conference,times}

\iclrfinalcopy 

\usepackage[utf8]{inputenc} 
\usepackage[T1]{fontenc}    
\usepackage{hyperref}       
\usepackage{url}            
\usepackage{booktabs}       
\usepackage{amsfonts}       
\usepackage{nicefrac}       
\usepackage{microtype}      
\usepackage{xcolor}         
\usepackage{amsmath}
\usepackage{graphicx}
\usepackage{float}
\usepackage{wrapfig}
\usepackage{subcaption}
\usepackage{algorithm}
\usepackage{algpseudocode}

\title{Dynamic Compression in Recurrent Networks}

\author{%
  Jyothish Pari\thanks{Correspondence: \texttt{jyop@mit.edu}. Code: \href{https://github.com/jyopari/dynamic-compression}{github.com/jyopari/dynamic-compression}.} \quad
  Ryan Bahlous-Boldi \quad
  Pulkit Agrawal \\
  Improbable AI Lab, Massachusetts Institute of Technology \\
}

\begin{document}

\maketitle

\begin{abstract}
Recurrent models process long contexts efficiently by compressing their history into a fixed-size state, but modern architectures typically do so in a single causal pass over the sequence. Each input must therefore be compressed before the model knows how it will later be used, forcing a limited state to compromise across possible future demands. We introduce \emph{dynamic compression}, which allows a recurrent model to selectively revisit past tokens and revise its fixed-size state through additional recurrent updates. The model need not preserve every part of the history at uniformly high fidelity in its recurrent state, because lower-fidelity information can be revisited from the retained raw sequence when it becomes relevant. We study this in a controlled setting where the model first learns multiple functions in-context and, later in the same sequence, encounters a series of few-shot tasks that each require it to identify and reuse one of those functions. A single-pass model must preserve every function at sufficient fidelity for any future task, whereas selective re-scanning allows the model to revisit and refine only the function currently needed. We find that dynamic compression substantially reduces the recurrent state required for accurate reuse and scales more favorably as the number of stored functions grows. These results demonstrate a computation--memory tradeoff in which recurrent models can spend more computation revisiting their history to make more effective use of a fixed-size state.
\end{abstract}

\section{Introduction}

Long-context inference can be viewed as a form of continual learning. As a sequence unfolds, a model must acquire information, retain it, and reuse it when later context makes that information relevant. We focus on recurrent models because of their favorable scaling with context length. For a prefix of length $t$, Transformers require $O(t)$ key--value memory and $O(t)$ attention computation per decoded token, whereas recurrent models require $O(1)$ recurrent-state memory and $O(1)$ recurrent computation per token with respect to $t$. This efficiency comes with a central constraint: how can a fixed-size recurrent state support an increasing amount of reusable information?

The causal recurrent architectures we study process context in a single left-to-right pass:
\begin{equation}
    S_t = f_\theta(S_{t-1}, x_t).
\end{equation}
Information is stored through updates to $S_t$ while the parameters $\theta$ remain fixed. Once $x_t$ has been processed, all future use of that token must occur through its contribution to $S_t$. The model must therefore decide how to represent $x_t$ before knowing which of its details future tasks will require.

For a task $\tau$ revealed after observing $x_{1:t}$, let $S_t^\star(\tau)$ denote an optimal state for solving it. Different tasks may require different allocations of the same finite state. The optimal state $S_t^\star(\tau_1)$ for one task may preserve one part of the history at high fidelity, while another task's optimal state $S_t^\star(\tau_2)$ may prioritize different information. A single-pass model, however, constructs $S_t$ before $\tau$ is known. It must therefore construct a single task-agnostic state that supports many possible task-specific states $S_t^\star(\tau)$. This forces the model to compromise across future task demands.

If information required to construct $S_t^\star(\tau)$ has been discarded from $S_t$, it cannot be recovered from the state alone. We therefore retain the observed prefix as a lossless record, making earlier compression decisions reversible. This changes our memory tradeoff. Our method uses $O(t)$ storage for the raw prefix while keeping the active recurrent state fixed-size. Once later context reveals what information matters, the model can revisit selected parts of this record and revise its recurrent state.

After observing $\tau$, the brute-force way to construct $S_t^\star(\tau)$ is to reread the entire prefix, requiring $O(t)$ additional recurrent updates for each task or query. We instead ask whether the model can select a smaller subset of previously observed positions $\mathcal{I}_\tau \subseteq \{1,\ldots,t\}$ and process the corresponding tokens again through the same recurrent update:

\begin{equation}
    S_t^{+} = f_\theta(S_t, x_{\mathcal{I}_\tau}),
\end{equation}

where we overload $f_\theta$ to denote sequential application over the selected tokens. The revised state $S_t^{+}$ replaces $S_t$ and subsequent processing continues from it. Figure~\ref{fig:method} illustrates this selective re-scanning process. Rather than searching the entire history at inference time, the model learns from the training distribution to predict which past positions are worth revisiting. In this work, we study a single selective re-scan before answering.

Because re-scanned tokens are processed through the persistent recurrence, choosing what to revisit also changes what is stored in the state. We call this capability \emph{dynamic compression}: the model can revise how its fixed-size state represents the past after later context reveals which information requires higher fidelity.

We study dynamic compression through a controlled continual function-reuse task. Within a single sequence, the model first observes several linear functions and later encounters few-shot tasks that require it to identify and reuse one of them. Identifying the relevant function requires only a coarse representation, whereas evaluating it on a new input requires substantially higher fidelity. A single-pass model must preserve every function accurately enough for possible future use; selective re-scanning allows the model to identify the relevant function and then revisit it when higher fidelity is required.

Because the locations of the tokens defining each function are known in our
controlled setting, we first evaluate selective re-scanning with oracle
supervision, providing an upper bound on the benefit of revisiting the correct
function. We then introduce a self-supervised procedure that derives re-scan
targets from the model's write strengths on a repeated context and trains a
dynamic model to predict them directly. Selective re-scanning substantially
reduces the recurrent-state capacity required for accurate function reuse and
scales more favorably than single-pass compression as the number of reusable
functions grows, exposing a tradeoff between recurrent-state capacity and
additional computation.

\section{Background}

\setlength{\abovedisplayskip}{8pt}
\setlength{\belowdisplayskip}{8pt}
\setlength{\abovedisplayshortskip}{6pt}
\setlength{\belowdisplayshortskip}{6pt}

A recurrent model compresses its history into a fixed-size state $\mathbf{S}_t$, updated
online and read to produce each output. We use \textbf{Gated DeltaNet} \citep{yang2025gated},
a linear-attention model \citep{pmlr-v119-katharopoulos20a} whose matrix-valued state
$\mathbf{S}_t \in \mathbb{R}^{d_v \times d_k}$ is updated by a gated delta rule
\citep{widrow1988adaptive, yang2024parallelizing}:
\[
\mathbf{S}_t
=
\mathbf{S}_{t-1}
\left(
\alpha_t
\left(
\mathbf{I}
-
\beta_t \mathbf{k}_t \mathbf{k}_t^\top
\right)
\right)
+
\beta_t \mathbf{v}_t \mathbf{k}_t^\top,
\qquad
\mathbf{o}_t = \mathbf{S}_t \mathbf{q}_t.
\]

The gates are computed from the layer input representation $x_t$ as
\[
\beta_t = \sigma(W_\beta x_t),
\qquad
\alpha_t =
\exp\!\left(
-A\,\operatorname{softplus}(W_\alpha x_t + b)
\right),
\]
where $\sigma$ denotes the sigmoid function and $A>0$ and $b$ are learnable
per-head parameters.

The key and value dimensions are $d_k = d_\text{head}$ and
$d_v = 2\,d_\text{head}$, where $d_\text{head}$ is the per-head dimension and
$2$ is a fixed value-expansion factor. The gate $\alpha_t \in (0,1)$ controls
global forgetting, while $\beta_t \in (0,1)$ controls the strength of the
key-specific delta update, determining how strongly the state is updated toward
$\mathbf{v}_t$ at $\mathbf{k}_t$. We therefore use $\beta_t$ as an
\emph{update-strength signal}. After the relevant task is known, we measure this
signal during an additional pass over the context and use it to derive
self-supervised targets for which tokens should be re-scanned.

The state is maintained per head, so the total recurrent state size (in elements) is
$n_\text{layer} \cdot n_\text{head} \cdot d_\text{head}^2 \cdot 2$. We fix
$n_\text{layer}=4$, $n_\text{head}=6$, and the value-expansion factor $2$
throughout, and vary only the head dimension $d_\text{head}$ to set the memory
budget.

\section{Task Description}

\begin{wrapfigure}{r}{0.5\linewidth}
    \vspace{-13pt}
    \centering
    \includegraphics[width=\linewidth]{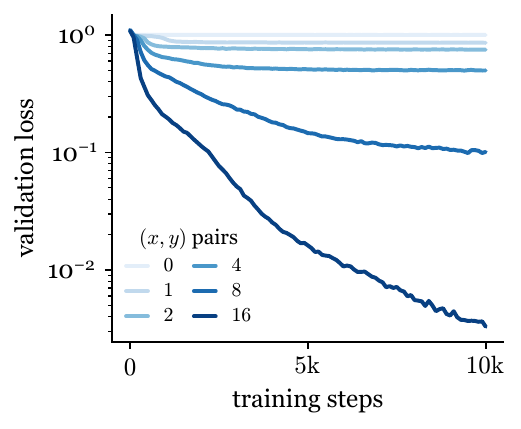}
    \vspace{-25pt}
    \caption{Validation loss during training, one curve per number of in-context pairs$(x_i, Ax_i)$. $x_i \sim \mathcal{N}(0,I)$, $A \sim \mathcal{N}(0,d^{-1})$, $d=8$, single basis $(K=1)$; the model predicts $Ax_{\mathrm{qry}}$. }
    \vspace{-15pt}
    \label{fig:context_scaling}
\end{wrapfigure}

The ability to internalize functions from context and reuse them on new inputs is a core requirement for continual learning. To study this concretely, we design a synthetic task that isolates functional reuse. The task is presented as a single continuous sequence with two phases.

We ground our choice of function class in a well-studied line of work showing that transformers can learn linear functions in-context, both theoretically and empirically \citep{akyurek2022learning, NEURIPS2022_c529dba0, pmlr-v202-von-oswald23a, fu2023transformers}. These works typically consider a scalar-output setting: given labeled pairs $(\mathbf{x}, w^\top \mathbf{x})$ with $\mathbf{x} \in \mathbb{R}^d$, the model must infer $w$ from context alone. We extend this to the vector-output setting, where both $\mathbf{x}$ and $\mathbf{y}$ are $d$-dimensional, so the function to learn is a full square matrix $\mathbf{A} \in \mathbb{R}^{d \times d}$. We first empirically verify that Gated DeltaNet can be trained from scratch to learn a single function $\mathbf{A}$ from context. The model stores the function in its recurrent state and correctly predicts $\mathbf{A}\mathbf{x}_{\mathrm{qry}}$ for a held-out query (Figure~\ref{fig:context_scaling}).

Having established that Gated DeltaNet can learn and reuse a single function from context, we next test whether a recurrent model can learn and reuse several functions within the same sequence. In the continual setting, the model must internalize $K$ different bases from a single sequence and later identify and reuse any of them on demand. We therefore pack all $K$ bases into one sequence and test whether the model can identify and apply the correct basis to subsequent queries.

We consider $K$ randomly drawn basis matrices $\mathbf{A}_1, \ldots, \mathbf{A}_K \in \mathbb{R}^{d \times d}$, with entries drawn i.i.d.\ from $\mathcal{N}(0,\, d^{-1})$. We use $d = 8$ throughout. The full input sequence is structured as follows:

\[
\underbrace{%
  \overbrace{t_1^{(1)} \cdots t_b^{(1)}}^{\mathbf{A}_1}
  \;\cdots\;
  \overbrace{t_1^{(K)} \cdots t_b^{(K)}}^{\mathbf{A}_K}
}_{\text{basis phase} \;(K \times b \text{ tokens})}
\;\Bigg|\;
\underbrace{%
  \overbrace{s_1^{(1)} \cdots s_f^{(1)}}^{\text{few-shot}}\; q^{(1)}
  \;\cdots\;
  \overbrace{s_1^{(T)} \cdots s_f^{(T)}}^{\text{few-shot}}\; q^{(T)}
}_{\text{query phase} \;(T \times (f+1) \text{ tokens})}
\]
\[
t_j^{(i)} = \bigl(\mathbf{x}_j,\; \mathbf{A}_i \mathbf{x}_j,\; 1,\; \mathbf{e}_i\bigr),
\qquad
s_j^{(t)} = \bigl(\mathbf{x}_j,\; \mathbf{A}_{i_t} \mathbf{x}_j,\; 1,\; \mathbf{0}\bigr),
\qquad
q^{(t)} = \bigl(\mathbf{x}_{\mathrm{qry}}^{(t)},\; \mathbf{0},\; 0,\; \mathbf{0}\bigr).
\]

Each token is a 4-tuple (input, output, binary flag, basis identifier); the flag indicates whether the output field contains a real $y$ value, and is $1$ for basis and few-shot tokens and $0$ for the query, giving input width $2d + 1 + K$. Basis tokens $t_j^{(i)}$ carry a one-hot identifier $\mathbf{e}_i \in \mathbb{R}^K$; few-shot and query tokens do not, so the model must infer the relevant basis from the examples alone.

\paragraph{Basis phase.}
For each basis $\mathbf{A}_i$, the model receives $b = 16$ labeled input-output pairs
\[
    \bigl(\mathbf{x}_j,\; \mathbf{A}_i\, \mathbf{x}_j\bigr), \qquad
    \mathbf{x}_j \overset{\mathrm{iid}}{\sim} \mathcal{N}(0, \mathbf{I}),
\]
with the binary flag set to $1$, indicating that the output field contains a real $y$ value. Each pair also carries a one-hot identifier $\mathbf{e}_i \in \mathbb{R}^K$ indicating the basis index. Since $b > d$, the pairs overdetermine $\mathbf{A}_i$. No prediction target is given during this phase.

\paragraph{Query phase.}
After the basis phase, the model processes $T$ query groups in sequence.
Each group is associated with a uniformly sampled basis
$i_q \in \{1,\ldots,K\}$ and consists of $f$ few-shot pairs
$(x_j, A_{i_q}x_j)$ with the binary flag set to $1$ and no basis identifier,
followed by a query input $x_{\mathrm{qry}}$ with the flag set to $0$.
The model is supervised on predicting $A_{i_q}x_{\mathrm{qry}}$.

The few-shot pairs are a search signal, not a learning signal: with $f < d$ they cannot recover $\mathbf{A}_{i_q}$ on their own, so the model must match them against its stored bases to identify the right one, then apply it to $\mathbf{x}_{\mathrm{qry}}$. Varying $K$ measures how difficulty scales with the number of functions the model must store and reuse.

Training from scratch and varying $K$ exposes the limit of static compression: $K=1$ is solved with a modest state, but matching that loss at $K=3$ requires roughly three orders of magnitude more state (${\sim}3\text{k}\to{\sim}3\text{M}$ elements). Packing more functions into one state under single-pass compression is prohibitively expensive.

Rather than scaling state size, can the model instead spend compute to compress past information more effectively within a smaller state?

\section{Selective Re-Scanning for Dynamic State Compression}
\label{sec:method}

\begin{wrapfigure}{r}{0.4\linewidth}
    \vspace{-10pt}
    \centering
    \includegraphics[width=\linewidth]{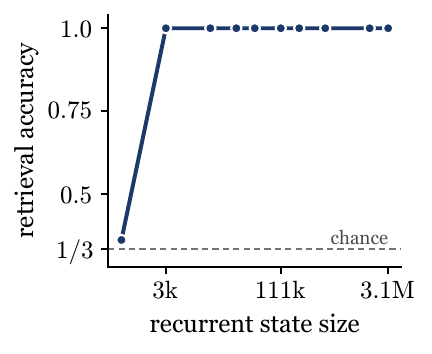}
    \vspace{-16pt}
    \caption{Validation accuracy of the selection head on the basis identification sub-task in isolation ($K=3$).}
    \label{fig:discrete_head}
\end{wrapfigure}

Retrieval over past context is well studied for transformers \citep{wu2022memorizingtransformers, mohtashami2023landmarkattentionrandomaccessinfinite}, where it is a pure read. Attention accesses past tokens without changing them. In an RNN, re-visiting a token is instead a new \emph{write} to the recurrent state, so re-scanning can change both what is stored and at what fidelity. This is dynamic compression: the model revises a token's contribution in light of later context, spending compute to raise fidelity where it matters most.

\begin{figure}[t]
    \centering
    \includegraphics[width=\linewidth]{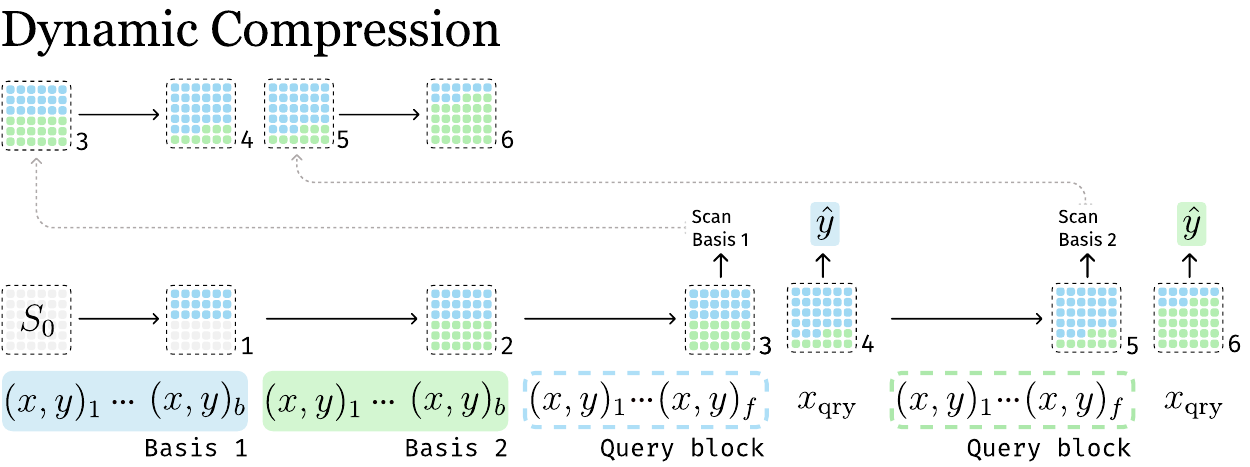}
     \caption{
    The selective re-scanning forward pass ($K=2$). In the basis phase, both bases are written into state sequentially. In the query phase, each query block first processes few-shot pairs to identify which basis is needed, re-scans that basis to sharpen its representation in state, then produces a prediction $\hat{y}$. Numbers at the lower right of each state denote the order of operations.
    }
    \label{fig:method}
\end{figure}

We augment Gated DeltaNet with two heads: a prediction head for the answer and a selection head that, at the end of each few-shot block, outputs a basis index $\hat{i} \in \{1, \ldots, K\}$. The model then re-scans all $b$ tokens of basis $\mathbf{A}_{\hat{i}}$, writing the selected function back into the state at higher fidelity. The following query reads from this revised state for prediction (Figure~\ref{fig:method}).

The two sub-tasks have different state requirements. Identifying the relevant basis needs only enough signal to distinguish among $K$ candidates. The selection head alone reaches near-perfect accuracy at approximately $3\text{k}$ state elements (Figure~\ref{fig:discrete_head}), far below the approximately $3\text{M}$ needed to store and apply three bases at once. The $K=1$ baseline fixes the state needed to apply a single basis accurately. A model that identifies the basis with a small state and then re-scans it at higher fidelity should therefore need far less state than storing all $K$ bases at full fidelity. For the main results (Figure~\ref{fig:combined_results}, Table~\ref{tab:dual_offline}) we supervise the selection head with the ground-truth index $i_q$ and re-scan the full $b$-token basis block as an oracle upper bound. Section~\ref{sec:learned} removes this oracle supervision and introduces a self-supervised method for learning what to re-scan.

\subsection{Toward Learned Selective Re-Scanning}
\label{sec:learned}

The oracle experiments assume supervision for which span to re-scan. We now introduce a self-supervised method that derives this supervision from the model's own write behavior when the context is repeated.
The key idea is to use the write strength $\beta$ on the additional pass as a signal for which past tokens the model would choose to update once the relevant task is known.

\paragraph{Step 1: extract a re-scan signal via repeated context.}
Inspired by \citet{arora2024justreadtwiceclosing}, we train a \emph{repeat model} on sequences where the full prefix is replayed before the query:

\[
\underbrace{%
  \overbrace{t_1^{(1)} \cdots t_b^{(1)}}^{\mathbf{A}_1}
  \;\cdots\;
  \overbrace{t_1^{(K)} \cdots t_b^{(K)}}^{\mathbf{A}_K}
}_{\text{basis phase}}
\;\Bigg|\;
\underbrace{s_1 \cdots s_f}_{\text{few-shot}}
\;\Bigg|\;
\underbrace{%
  \overbrace{t_1^{(1)} \cdots t_b^{(1)}}^{\mathbf{A}_1}
  \;\cdots\;
  \overbrace{t_1^{(K)} \cdots t_b^{(K)}}^{\mathbf{A}_K}
  \;\overbrace{s_1 \cdots s_f}^{\text{few-shot}}
}_{\text{repeat}}
\;\Bigg|\;
q
\]

On the repeated prefix, the model has already observed the few-shot block, so its
writes can depend on which past information is now relevant. We therefore treat
the write strength $\beta_t$ on this additional pass as a candidate re-scan
signal. In our experiments, the final-layer $\beta$ pattern varies systematically
with the queried basis (Figure~\ref{fig:betas}, top; additional examples in
Appendix~\ref{fig:betas_appendix}), motivating the codebook construction below.

\paragraph{Step 2: convert $\beta$ into a re-scan codebook.}
We collect the final-layer $\beta$ values over the 52-token repeat region from
many sequences, giving one write-strength pattern per sequence. We cluster these
patterns with $k$-means and choose $C=3$, after which increasing the number of
clusters yields little further reduction in clustering loss. Each cluster
centroid identifies past positions that tend to receive high write strength. We
convert each centroid into a fixed-length re-scan choice containing $n=21$
positions. Together, these choices form a small codebook that the dynamic model
learns to predict. We describe the layer selection and the construction of the
fixed re-scan length in Appendix~\ref{app:betas}. This codebook is specific to our synthetic setting; more general re-scan
parameterizations remain open, such as letting the model place its own
``bookmarks'' in the context, use tools to search the history, or navigate the
context through learned relative movements
\citep{zaremba2015reinforcement}.

\begin{figure}[t]
    \centering
    \includegraphics[width=\linewidth]{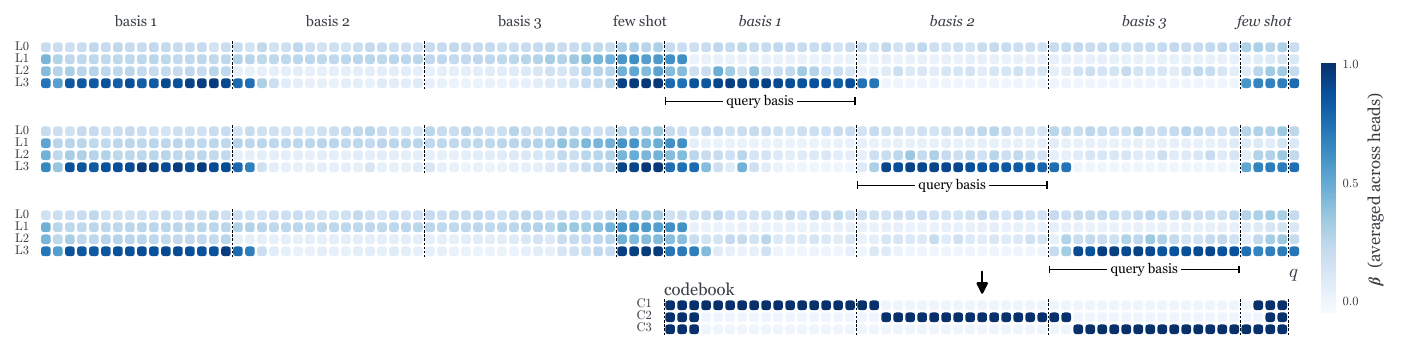}
    \caption{
        \textbf{Top:} Write strength $\beta$ (averaged across heads) across layers and token positions for three repeat-model sequences, each containing $K=3$ bases and one query group.
        \textbf{Bottom:} A re-scan codebook constructed from the $\beta$ patterns of many such sequences. The codebook compresses these patterns into a small set of representative re-scan choices for the dynamic model to predict.
        }
    \label{fig:betas}
\end{figure}

\paragraph{Step 3: train the dynamic model.}
Using the frozen repeat model and codebook, we generate training sequences of the
form
\[
    [\,\text{basis} \mid \text{few-shot} \mid \text{re-scan} \mid \text{query}\,],
\]
where each sequence is assigned to its nearest code and the corresponding
$n=21$ prefix tokens form the re-scan block. The dynamic model is trained to
predict the re-scan code after the few-shot block and the query output after
processing the selected tokens.

At inference, no repeated context is needed: the model processes the prefix,
predicts a code, re-scans the corresponding tokens, and produces its answer. Table~\ref{tab:dual_offline} compares all four model families by their best validation MSE (mean $\pm$ std over 3 seeds): single-pass baseline, repeat model, oracle dynamic (ground-truth supervision), and codebook dynamic (unsupervised). The codebook dynamic sits well below the single-pass baseline and partway toward the oracle and repeat upper bounds, demonstrating that the label-free $\beta$ signal is sufficient to learn an effective re-scanning policy.

\begin{figure}[H]
    \centering
    \includegraphics[width=\linewidth]{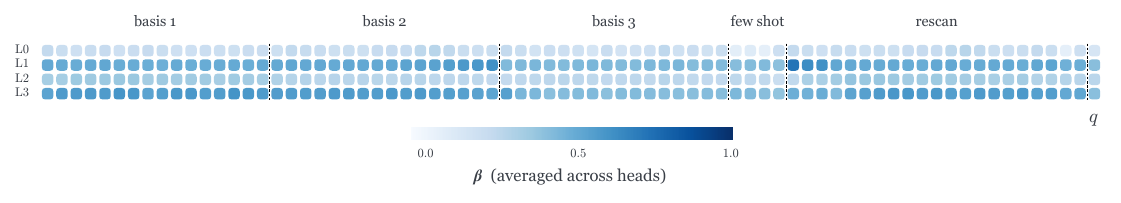}
    \caption{Write strength $\beta$ of the trained codebook dynamic model on its autoregressive re-scan path, for one example ($K=3$). On the first pass, the model writes the three bases relatively evenly; after selecting a basis, it re-scans that basis and updates it more strongly.
    }
    \label{fig:dual_betas}
\end{figure}

\paragraph{Dynamic re-scanning changes how the model compresses.}
The repeat and dynamic models exhibit different write patterns (Figures~\ref{fig:dual_betas}
and~\ref{fig:betas}). The repeat model can defer high-strength updates until the context
is replayed, whereas the dynamic model must retain enough information on the first pass to decide
what to re-scan. After selecting a region, it can then refine that information through additional
recurrent updates.

\section{Results}

\begin{figure}[t]
    \centering
    \begin{minipage}[t]{0.49\linewidth}
        \centering
        \includegraphics[width=\linewidth]{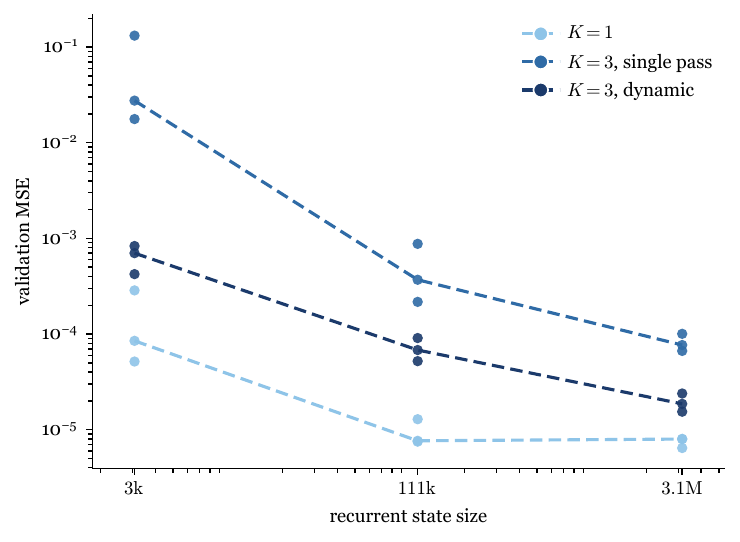}
    \end{minipage}
    \hfill
    \begin{minipage}[t]{0.49\linewidth}
        \centering
        \includegraphics[width=\linewidth]{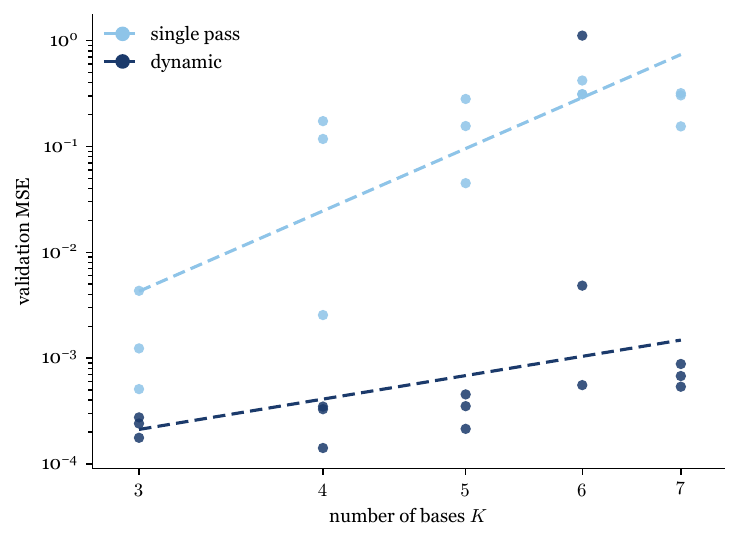}
    \end{minipage}
    \caption{
        \textbf{Left:} Validation MSE versus recurrent state size for $K=1$ single-pass,
        $K=3$ single-pass, and $K=3$ oracle dynamic re-scanning. Points show individual
        seeds and dashed lines connect the median across seeds.
        \textbf{Right:} Validation MSE versus the number of bases $K$ at
        $d_{\mathrm{head}}=12$ for single-pass and oracle dynamic re-scanning, with
        power-law fits to the median across seeds.
        }
    \label{fig:combined_results}
    \label{fig:k_sweep}
\end{figure}

Oracle dynamic re-scanning consistently outperforms single-pass compression
across recurrent state sizes (Figure~\ref{fig:combined_results}), with the
largest gains under tight memory budgets. At $K=3$, dynamic re-scanning with a
$111$k-element state reaches lower error than the single-pass model with a
$3.1$M-element state.

At fixed state size, the advantage grows as the number of functions increases.
Both methods degrade with $K$, but single-pass error grows substantially faster,
showing that selective re-scanning scales more favorably as more functions must
be retained for future reuse. One dynamic run at $K=6$ failed to optimize and is
shown at the single-pass level.

\begin{table}[h]
    \centering
    \caption{
    Best validation MSE (mean $\pm$ std over 3 seeds, in units of $10^{-3}$)
    for the four model families at $d_\text{head}=16$, $50$k training steps,
    and a single query group. The single-pass baseline mean is inflated by one
    high-variance seed; its median is $17.4$.
    }
    \label{tab:dual_offline}
    \begin{tabular}{lc}
    \toprule
    Method & Validation MSE ($\times 10^{-3}$) \\
    \midrule
    Single-pass baseline & $67.06 \pm 95.99$ \\
    Repeat & $1.38 \pm 0.44$ \\
    Oracle dynamic & $2.30 \pm 2.44$ \\
    Codebook dynamic & $4.91 \pm 1.53$ \\
    \bottomrule
    \end{tabular}
\end{table}

Table~\ref{tab:dual_offline} evaluates the learned re-scanning mechanism.
The codebook dynamic model substantially improves over the single-pass baseline
and closes part of the gap to oracle dynamic re-scanning, showing that useful
re-scan decisions can be learned without oracle re-scan supervision.

More broadly, lifelong settings may require models to accumulate and reuse an
increasing number of skills learned earlier in context. Rather than forcing a
fixed recurrent state to preserve all of this information at uniformly high
fidelity, retaining the raw context and selectively revisiting it allows the
model to dynamically recompress the past around what is relevant to the current
task. Our results suggest that this may provide a more scalable way to support
continual reuse as the amount of previously learned information grows.

\section{Related Work}

\paragraph{Modern RNNs and their limitations.}
Linear attention admits a recurrent formulation in which past context is compressed into a fixed-size state \citep{katharopoulos2020transformers}. Building on this formulation, \citet{schlag2021linear} introduce DeltaNet, which replaces additive fast-weight updates with the delta rule \citep{widrow1988adaptive}, and subsequent work develop increasingly efficient and expressive versions  \citep{yang2024parallelizing, yang2025gated}. Titans and Atlas further introduce richer test-time memory updates \citep{behrouz2026titans,behrouz2025atlas}. In this work, we use Gated DeltaNet, which augments the delta update with learned gates \citep{yang2025gated}. Despite these advances, recurrent models retain two fundamental constraints: their recurrent state has finite capacity, and standard inference processes the context in a single left-to-right pass. \emph{Just Read Twice} addresses the latter by rereading the context \citep{arora2024justreadtwiceclosing}. We build on this idea by using an additional pass not only to recover information, but also to derive supervision for which parts of the context should be revisited. Other work addresses the finite-state bottleneck by allowing memory to grow with context, including Memory Caching and Log-Linear Attention \citep{behrouz2026memory,guo2026log}. Our approach instead keeps the recurrent state fixed-size while retaining the raw tokens as a lossless record that the model can selectively revisit to revise its state for the current task.

\paragraph{Selective access to past context.}
Several works learn which parts of a long context deserve additional computation. Sparse-attention methods such as Native Sparse Attention and DeepSeek-V4's Compressed Sparse Attention restrict attention to a selected subset of past context \citep{yuan2025native,xu2026deepseek}. Self-Guided Test-Time Training similarly selects question-relevant spans, but uses them to update model parameters \citep{zhu2026self}. Resona augments recurrent models with retrieval from the original context and integrates retrieved chunks through cross-attention \citep{wang2025resona}. Concurrent work, HOLA, augments Gated DeltaNet with an exact KV cache and independently uses a retention criterion based on the delta-rule update magnitude $\beta\lVert e\rVert$ to decide which tokens to retain \citep{cui2026hippocampus}. In contrast, our method selectively reprocesses raw past tokens through the same recurrence, using task-informed writes on an additional pass to supervise what should be revisited and thereby revise the fixed-size recurrent state.

\section{Discussion and Limitations}

Continual learning with a recurrent network is fundamentally a compression problem: a growing context must be represented within a fixed-size state, and the quality of that compression determines what the model can later reuse. Our results show that selective re-scanning allows the model to improve this compression by spending additional computation where higher fidelity is needed, substantially reducing the state required to reuse multiple functions.

Dynamic compression creates a two-level memory hierarchy: the raw context provides lossless access to the past, while the recurrent state serves as a compact working memory that can be revised as new tasks reveal which information matters. Rather than fixing the compression of each token when it is first observed, the model can revisit and rewrite past information as its relevance becomes known. More broadly, this raises a question beyond memory efficiency: if a model can learn to compress its history more effectively, can scaling this capability also improve generalization? 

We study only a controlled synthetic setting here, and extending dynamic compression to natural-data pretraining raises several open problems. In particular, we must determine how to parameterize the space of possible re-scans and how to generate supervision for where to revisit when tasks are not explicitly delineated in the data. Post-training may provide a complementary setting for studying dynamic compression at scale, where task structure is more controlled and reinforcement learning can be used to explore re-scan decisions directly.

\section*{Acknowledgements}
We want to express our gratitude to Yoon Kim, Samuel J. Gershman, Freda Shi, Shivam Duggal, Lucas Torroba-Hennigen, Han Guo, Yikang Shen, Mayank Mishra, Zhang-Wei Hong, Nitish Dashora, Akarsh Kumar and members of the Improbable AI lab for the helpful discussion on the paper. Research was sponsored by the Army Research Office and was accomplished under Grant Number W911NF-23-1-0277. The views and conclusions contained in this document are those of the authors and should not be interpreted as representing the official policies, either expressed or implied, of the Army Research Office or the U.S. Government. The U.S. Government is authorized to reproduce and distribute reprints for Government purposes notwithstanding any copyright notation herein.

\bibliographystyle{iclr2026_conference}
\bibliography{bibliography}

\appendix

\section{Additional write-strength ($\beta$) visualisations}
\label{fig:betas_appendix}

\begin{figure}[H]
    \centering
    \includegraphics[width=\linewidth]{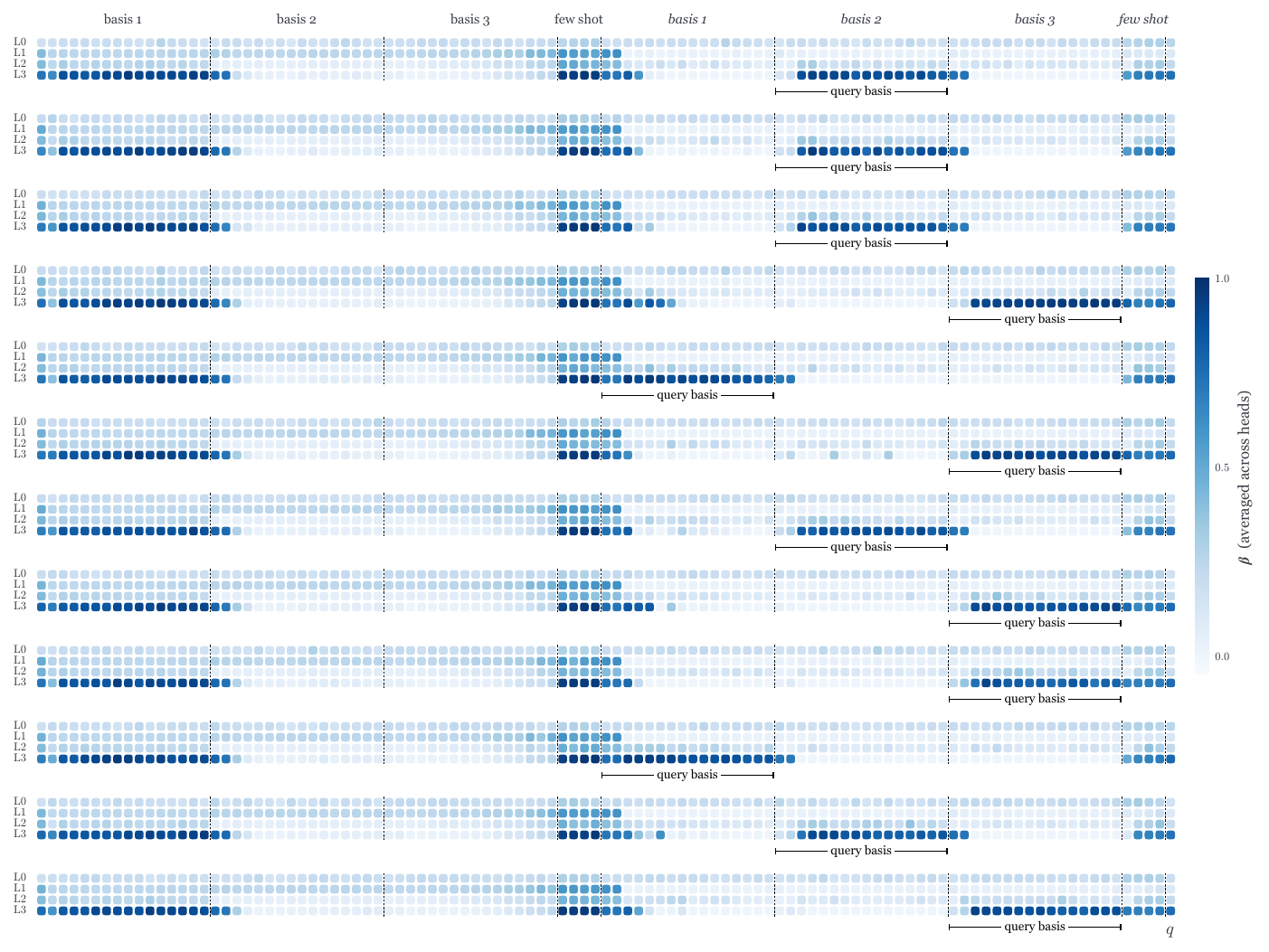}
    \caption{Write strength $\beta$ for a random sample of repeat-model sequences ($K=3$); the \emph{query basis} span rule marks the repeated block of the matched basis.}
\end{figure}

\begin{figure}[H]
    \centering
    \includegraphics[width=\linewidth]{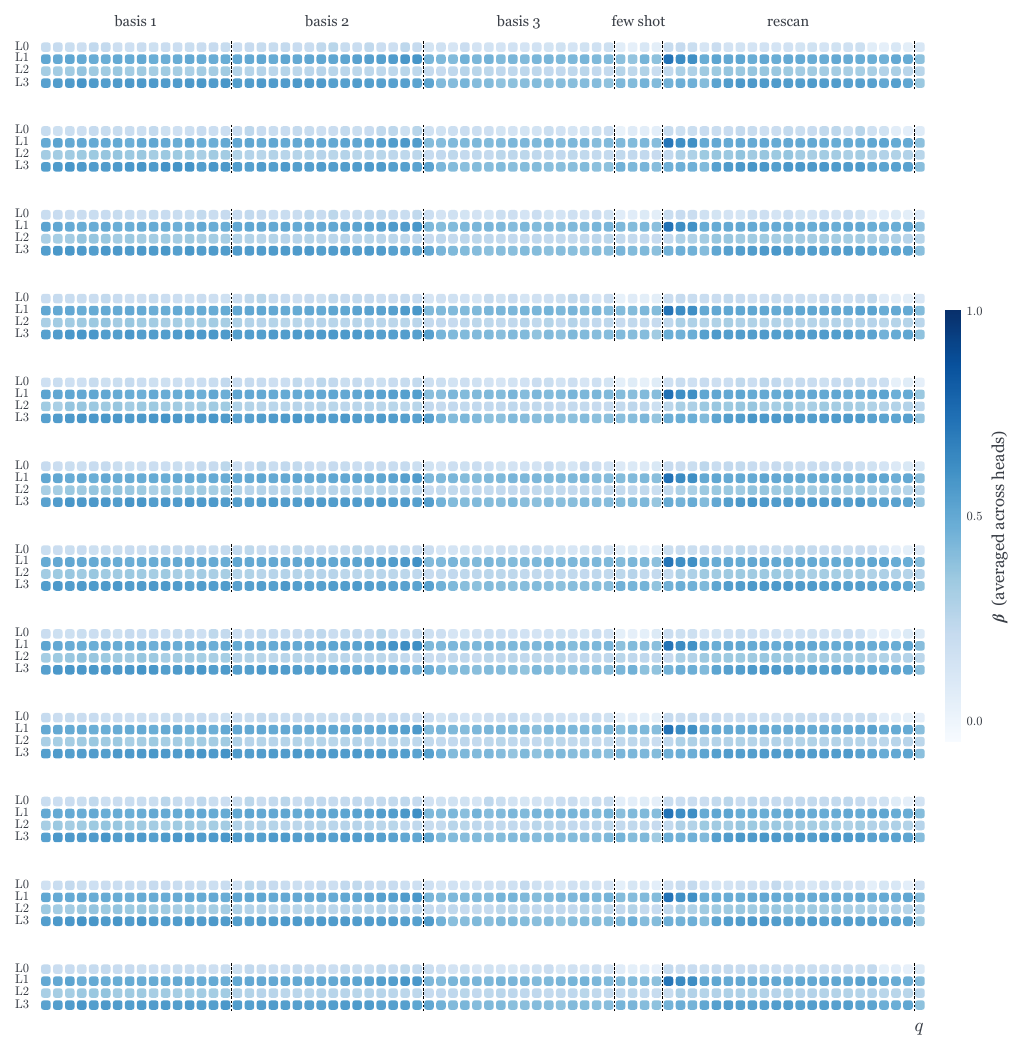}
    \caption{Dynamic-model $\beta$ (as in Figure~\ref{fig:dual_betas}) for a random sample of sequences.}
\end{figure}

\section{Experimental details}
All models are Gated DeltaNet with $4$ layers, $6$ heads, embedding dimension $256$, and value expansion $2$; the head dimension $d_\text{head}$ sets the recurrent state size $4 \cdot 6 \cdot d_\text{head}^2 \cdot 2$. Training uses AdamW ($\beta_1{=}0.9$, $\beta_2{=}0.95$, weight decay $10^{-2}$), a constant learning rate of $10^{-4}$, and gradient-norm clipping at $1.0$, with no learning-rate schedule; validation is evaluated $100$ times over training. The underlying task is the in-context matrix-regression task of the main text with token dimension $d=8$. Each subsection below gives the settings specific to one figure, the conditions and seeds compared, and how the reported quantity is computed.
\subsection{State-size figure (Figure~\ref{fig:combined_results})}
\label{app:main}
Head dimensions $d_\text{head} \in \{8, 48, 256\}$ give state sizes $3072$, $110{,}592$, and
$3{,}145{,}728$ (i.e.\ ${\approx}3\text{k}$, $111\text{k}$, $3.1\text{M}$). Each sequence has
$M=16$ demonstration pairs per basis, $F=4$ few-shot pairs, and $8$ query positions; training
is $150{,}000$ steps at batch size $512$ with $10{,}000$ held-out validation sequences.
Three conditions are compared, each at three seeds: single basis ($K=1$), single pass
($K=3$), and dynamic re-scanning ($K=3$). The first two use the plain prediction head and are
scored by validation MSE; dynamic re-scanning adds the selection head and is scored by
its autoregressive re-scan MSE (predict a basis, re-scan it, then answer the query). Reported:
for each run, the minimum validation MSE over training; every seed is plotted as a point, and
dashed lines connect the per-state-size medians.

\subsection{Scaling with the number of bases (Figure~\ref{fig:k_sweep})}
\label{app:scaling}

The head dimension is fixed at $d_\text{head}=12$ (state size $6912$) for every point. Each
sequence has $M=16$ demonstration pairs per basis, $F=4$ few-shot pairs, and $8$ query
positions; training is $150{,}000$ steps at batch size $512$ with $10{,}000$ validation
sequences. Two conditions---single pass and dynamic re-scanning---are compared at
$K \in \{3,4,5,6,7\}$ with three seeds per setting. For each run, we report the minimum
validation MSE over training. Every seed is plotted individually, and the power-law fit is
computed from the median across seeds at each $K$. The fitted exponents are
${\approx}6.1$ for single pass and ${\approx}2.3$ for dynamic re-scanning. One dynamic
seed at $K=6$ failed to optimize and sits at the single-pass level.
\subsection{Offline dynamic-model comparison (Table~\ref{tab:dual_offline})}
\label{app:dual_offline}
All families use head dimension $16$ (state size $12{,}288$), $K=3$ bases, $M=16$ demonstration
pairs per basis, $F=4$ few-shot pairs, and a single query group per sequence; training is
$50{,}000$ steps at batch size $1024$ with $50{,}000$ validation sequences. Four families are
compared, each at three seeds:
\begin{itemize}
    \item \textbf{single-pass baseline} --- reads the prefix once;
    \item \textbf{repeat} --- re-reads the entire prefix before the query;
    \item \textbf{oracle dynamic} --- re-scans the ground-truth basis block, with the selection head supervised by the true basis index;
    \item \textbf{codebook dynamic} --- re-scans the tokens chosen by the $\beta$-derived codebook (Appendix~\ref{app:betas}), using no ground-truth basis labels.
\end{itemize}
The baseline and repeat families are scored by validation MSE; the two dynamic families by the
autoregressive re-scan MSE. The codebook dynamic is trained on a fixed, pre-generated set of
$5 \times 10^{7}$ sequences of the form [basis $|$ few-shot $|$ re-scan $|$ query], where the
re-scan block is the codebook's selection for that sequence; each seed uses an independently
generated dataset. Reported: for each run, the minimum validation MSE over training; the table
gives the mean $\pm$ sample standard deviation across the three seeds, in units of $10^{-3}$.
The single-pass baseline mean is inflated by one high-variance seed (its median is
$17.4 \times 10^{-3}$).

\subsection{Selection-head retrieval accuracy (Figure~\ref{fig:discrete_head})}
\label{app:discrete}
Here the model has only a selection head that predicts the query basis index ($K=3$, so chance
is $1/3$), trained by cross-entropy. The head dimension is swept over
$d_\text{head} \in \{4, 8, 16, 24, 32, 48, 64, 96, 192, 256\}$, giving state sizes from $768$
to ${\approx}3.1\text{M}$; each sequence has $M=16$ demonstration pairs per basis, $F=4$
few-shot pairs, and $8$ query positions; training is $20{,}000$ steps at batch size $512$ with
$128$ validation sequences, one seed per head dimension. Reported: for each run, the maximum
validation accuracy over training, plotted against state size.

\subsection{Context scaling (Figure~\ref{fig:context_scaling})}
\label{app:context}
This is a single-basis task ($K=1$) at head dimension $16$. Each sequence is $M$ demonstration
pairs $(\mathbf{x}, \mathbf{A}\mathbf{x})$ followed by $8$ query positions, with no few-shot
block ($F=0$); $M$ is swept over $\{0, 1, 2, 4, 8, 16\}$. Training is $10{,}000$ steps at batch
size $128$ with $256$ validation sequences, one run per value of $M$. Reported: each run's
validation-loss curve is plotted directly against training step (log-$y$), with no
aggregation.

\subsection{Write-strength ($\beta$) analysis and codebook
(Figures~\ref{fig:betas}, \ref{fig:dual_betas})}
\label{app:betas}

These figures visualize the delta-rule write strength $\beta \in (0,1)$,
recorded per layer, token, and head during the forward pass and averaged across
heads. Both source models use the shared architecture at head dimension $16$,
trained for $50{,}000$ steps at batch size $1024$ (as in
Appendix~\ref{app:dual_offline}): the repeat model for
Figure~\ref{fig:betas} and the codebook dynamic model for
Figure~\ref{fig:dual_betas}.

\paragraph{Repeat-model $\beta$ and codebook.}
We forward repeat-format sequences
(basis $|$ few-shot $|$ repeated prefix $|$ query) and record $\beta$ over the
repeated region. We inspect the head-averaged $\beta$ patterns across layers and
use the final layer, whose write pattern most clearly aligns with the queried
basis in our experiments. More generally, write strengths could instead be
aggregated across layers, for example by summing $\beta$, but we do not explore
this here.

Each sequence is represented by a $52$-dimensional vector of final-layer
$\beta$ values over the repeated region. We cluster these vectors with
$k$-means, choosing the number of codes from the elbow in the clustering loss,
which occurs at $C=3$.

Each centroid defines a re-scan pattern over the $52$ prefix positions. We
threshold each centroid at $\beta_{\max}/2$ and set the common re-scan length
$n$ to the smallest number of above-threshold positions across centroids,
giving $n=21$. Each code then selects its top-$n$ positions by centroid value.
This gives every code an equal-length re-scan block while ensuring that all
selected positions exceed the threshold.

\paragraph{Dynamic-model $\beta$.}
The trained codebook dynamic model is run along its inference path---read the
prefix, predict a code, re-scan that code's tokens, then answer---and $\beta$ is
recorded over the resulting sequence
(basis $|$ few-shot $|$ re-scan $|$ query).

\end{document}